\documentclass[letterpaper, 10 pt, conference, twocolumn]{ieeeconf}

\IEEEoverridecommandlockouts 
\usepackage{amsmath} 
\usepackage{amssymb} 

\usepackage[dvipdfmx]{graphicx}
\usepackage{multirow}
\usepackage{cite}
\usepackage{url}

\usepackage{mleftright}

\usepackage[%
 dvipdfmx,%
 setpagesize=false,%
 bookmarks=false,%
 bookmarksdepth=tocdepth,%
 bookmarksnumbered=false,%
 colorlinks=true,%
 linkcolor=black,
 urlcolor=black,
 pdftitle={},%
 pdfsubject={},%
 pdfauthor={},%
 pdfkeywords={}%
]{hyperref}

\usepackage{color}
\usepackage[table]{xcolor}
\usepackage{tabularx}
\newcolumntype{Y}{&gt;{\centering\arraybackslash}X} %

\newcommand{\E}{\mathbb{E}}

\usepackage[ruled,vlined]{algorithm2e}
\usepackage{algorithmic}

\usepackage{makecell}

\newcommand{\bhline}[1]{\noalign{\hrule height #1}}

\title{\LARGE \bf
The Impact of Overall Optimization on Warehouse Automation
}

\author{Hiroshi Yoshitake$^{1}$ and Pieter Abbeel$^{2}$ 
\thanks{$^{1}$H. Yoshitake is with R\&D Group, Hitachi Ltd., Kokubunji-shi, Tokyo, Japan {\tt\small hiroshi.yoshitake.nt@hitachi.com}, 
$^{2}$P. Abbeel is with Department of EECS, University of California, Berkeley, CA, USA
{\tt\small pabbeel@cs.berkeley.edu}
}
}

\begin{document}

\maketitle
\thispagestyle{empty}
\pagestyle{empty}

\begin{abstract}
In this study, we propose a novel approach for investigating optimization performance by flexible robot coordination in automated warehouses with multi-agent reinforcement learning (MARL)-based control. 
Automated systems using robots are expected to achieve efficient operations compared with manual systems in terms of overall optimization performance. 
However, the impact of overall optimization on performance remains unclear in most automated systems due to a lack of suitable control methods. 
Thus, we proposed a centralized training-and-decentralized execution MARL framework as a practical overall optimization control method. 
In the proposed framework, we also proposed a single shared critic, trained with global states and rewards, applicable to a case in which heterogeneous agents make decisions asynchronously. 
Our proposed MARL framework was applied to the task selection of material handling equipment through automated order picking simulation, and its performance was evaluated to determine how far overall optimization outperforms partial optimization by comparing it with other MARL frameworks and rule-based control methods. 
\end{abstract}

\section{INTRODUCTION}
In recent years, industrial automation has rapidly advanced with the active installation of robots. 
Many industrial sites, such as logistics warehouses and production lines, are experiencing labor shortages. 
They also require highly efficient and long-hour operations to deal with bulk orders promoted by E-commerce \cite{custodio20}. 
To address these issues, manual operations are automated by replacing human laborers with industrial robots. 
The installation of industrial robots increases at an annual rate of $\sim$10\%, and robotic automation helps compensate for the workforce shortage \cite{ifr21}. 
More efficient automation techniques are also required to improve the effectiveness of robot installation. 

One of the reasons why the installation of automated systems is expected to increase efficiency is the possibility of further efficient operations based on overall optimization. 
In conventional manual operations, detailed control and modeling are challenging due to the several uncertainties associated with human actions and decision making. 
If automation progresses using industrial robots, their predictable and deterministic behaviors will increase the control accuracy in operations. 
By constructing such highly accurate control schemes in various processes, there is a prospect of optimizing the entire system. 
However, most of the previous studies have only addressed partial optimization for limited cases, such as a part of an automated system or material handling equipment (MHE) for a certain process \cite{azadeh19}. 
It is unclear how much the overall optimization of multiple processes will affect the automated system. 

Multi-agent reinforcement learning (MARL) can provide practical overall optimization for controlling industrial automated systems. 
A control method of robot coordination for the overall optimization has not yet been established. 
However, centralized training with decentralized execution (CTDE), one of the major MARL frameworks in which agents make decisions in a decentralized manner and learn the coordination using centralized estimators \cite{nguyen20}, would be a solution for it:  
cyber-physical modeling enables the training of reinforcement learning (RL) agents centrally \cite{omniverse}. 
Furthermore, a decentralized RL policy would be reasonable for practical implementation: 
a centralized controller (e.g., warehouse management system) would not control thousands of robots one by one in terms of system load \cite{Wurman08}. 
In this case, it is more natural that the robot fleet is controlled by a decentralized sub-system operating the corresponding process. 
Furthermore, there is currently no unified controller that can handle various types of MHE and robots released from different vendors. 

In this study, we investigate the impact of overall optimization on automated systems, particularly warehouse automation using different MARL frameworks. 
There are two main challenges for the MARL-based approach in robot coordination, which is essential for the overall optimization of industrial automation. 
The first one is a long-horizon task under extremely sparse reward settings: evaluation indices of system performance such as task completion time (makespan) and productivity can be estimated only at the terminal state when all tasks are completed. 
The second one is the asynchronous decision making of heterogeneous agents: when various types of robots execute tasks, they do not make decisions under the synchronous settings that major MARL algorithms assume. 
We, therefore, examine an effective MARL framework that maximizes the performance of automated systems under these challenges by introducing a simplified automated warehouse simulator. 

The purpose of this study is to answer the following research questions ({\bf RQ}s).
\begin{itemize}
\item {\bf RQ1}: How can MARL agents acquire coordinated behaviors in industrial automated environments?
\item {\bf RQ2}: Which coordination condition is important for overall optimization?
\item {\bf RQ3}: What advantages does overall optimization bring to automated systems compared with partial optimization?
\end{itemize}
The contributions of this study are as follows.
\begin{itemize}
\item Proposal of a CTDE-based practical MARL algorithm applicable to industrial automated environments. 
\item Building a simulation environment of an automated order picking system\footnote[1]{The source code is available at \url{https://github.com/16444take/aope-sim.git}}.
\item Clarifying effective coordination conditions and the impact of overall optimization when using the proposed MARL algorithm and simulation environment. 
\end{itemize}

The remainder of this paper is organized as follows. In \S II, we explain related works. We provide a detailed description of the proposed MARL frameworks and algorithms in \S III. In \S IV, we evaluate the proposed methods. We conclude the study and discuss future work in \S V.

\section{Related Works}

\subsection{Optimization of Logistics Warehouse Operation}

The operational purpose of logistics warehouses is to collect and ship inventory items according to orders received from customers \cite{Bartholdi17}. 
To achieve this order fulfillment, warehouse operations consist of several processes: receiving, inventory control, order picking, inspection, packing, and shipping. 
Each process also includes the workforce such as human laborers and MHE, as well as their tasks to handle the ordered items. 
Optimizing these processes helps maintain efficient warehouse operations, and thus, minimizing the makespan and maximizing the system throughput are key research topics for warehouse optimization \cite{jaghbeer20}. 
Because warehouse automation has made operation control more accurate by replacing human laborers with robots, optimization has become even more crucial for efficiency \cite{zhen22}. 

Maximizing the efficiency of warehouse operations requires control technology that comprehensively optimizes all related processes. 
Despite the widely recognized need for such technology, previous studies have been limited to partial optimization within a certain process such as optimal routing and order batching for manual/robotized order picking and optimal storage assignment based on order frequency for inventory control \cite{azadeh19}. 
Therefore, an overall optimization method that considers the operation of multiple processes has not yet been established. 
In this study, we introduce a state-of-the-art (SOTA) MARL algorithm as an overall optimization method for warehouse operations.

\subsection{Multi-Agent Reinforcement Learning}

MARL, an extension of the RL framework, where several agents execute different tasks, is extensively studied for acquiring the optimal policies of agents in a multi-agent system (MAS). 
Agents need to learn by considering not only information about their local environments but also other learning agents. 
Many MARL frameworks have been proposed depending on various conditions such as operational settings, where agents are completely controlled by a central unit or operate autonomously in decentralized settings, and situations in which the tasks of agents are stationary or non-stationary \cite{nguyen20}. 

MARL-based control with a decentralized policy would be reasonable for automated systems due to its scalability. 
The simplest framework of such control is independent learning (IL), where each agent is trained in the same manner as in single-agent RL: an agent independently learns the policy from its local observations and behaviors \cite{tan93} even though it suffers from instability of the training environment caused by the policy updates of other agents. 
Recent MARL studies have focused on the CTDE framework where decentralized policies are trained by centralized critics that estimate the contributions of all agents. 
One of the most widely referenced CTDE frameworks is the multi-agent deep deterministic policy gradient (MADDPG), which uses joint critics of the states and actions of all agents\cite{lowe17}. 
We apply both IL and CTDE frameworks to automated warehouse operations and clarify how much they can contribute to the optimization.

\subsection{Applications of MARL in Industrial Automation}

Applications of MARL are frequently studied in many industrial fields such as manufacturing, logistics, networking, and automotive \cite{pulikottil21, canese21}. 
Most industrial systems require intelligent controllers that create operation schedules or send instructions to the control object to maximize performance. 
RL provides reasonable solutions to these problems compared with other optimization techniques \cite{bello16} and has the advantage of flexibility: 
RL agents are trained by experiencing various situations in a system, and the trained policy can output optimal actions for any states. 
Because unexpected variable factors such as noises and disturbances tend to occur in industrial systems, the control of agents, resource allocation, and task planning is expected to be flexible in response. 
The goal of MARL research in the industrial field is to incorporate the above RL features into systems consisting of several autonomous agents. 

One of the most popular MARL applications is the control of MHE used for automating order picking in logistics warehouses and factories \cite{shen21}. 
Order picking is an operation, where ordered items are collected for shipping destinations from warehouse storage. 
The transfer of items during order picking has been recently automated by MHE consisting of several automated guided vehicles (AGVs) \cite{azadeh19}. 
The task allocation or path planning of AGVs has been successfully executed by introducing recent CDTE algorithms \cite{li21,shen21}. 
However, such MARL-based control has only been applied to the process of item transfer by homogeneous robots, whereas order picking includes other processes involving heterogeneous robots such as picking and placing items by picking robots and subsequent transfer of items by conveyors. 
Few previous studies have addressed more complicated cases that allow training heterogeneous agents with asynchronous decision making \cite{Xiao20,wang21,xiao22}; however, they used the MADDPG (i.e., off-policy MARL algorithm), making it difficult to apply them to the sparse reward setting, one of the typical features in industrial automation. 
In this study, we develop an MARL framework that can be applied to several processes with heterogeneous agents.

\section{METHODOLOGY}

\subsection{Problem Settings}

Although the operational status of automated systems can be disturbed by noise and stochastic factors, it depends on the decision making for controlling MHE and robots that perform the tasks in each process. 
By regarding the controller of MHE and robots as an autonomous agent, state changes in an automated system can be described as a Markov decision process: $\langle \mathcal{S}, \mathcal{A}, \mathcal{P}_{\mathrm{T}}, \mathcal{R} \rangle$, where $s\in\mathcal{S}$ denotes a set of system states, $a\in\mathcal{A}$ denotes a set of agents' actions, $\mathcal{P}_{\mathrm{T}}: \mathcal{S}\times\mathcal{A}\times\mathcal{S}\rightarrow [0,1]$ denotes the state transition probability organized by the system operations, and $r\in\mathcal{R}:\mathcal{S}\times\mathcal{A}\times\mathcal{S} \rightarrow \mathbb{R}$ denotes the reward function, respectively. 
Assuming a situation, where the controller makes decisions solely based on the states of its corresponding process, we further study the decentralized partially observable MDP as $\langle \mathcal{S}, \{\mathcal{A}^{i}\}, \mathcal{O}, \mathcal{P}_{\mathrm{T}}, \mathcal{R}, N \rangle$, where $o^{i} \in \mathcal{O}$ denotes the local observation for agent $i\in N$ at global state $s$, and $a^{i} \in \mathcal{A}^{i}$ denotes an action set of each agent \cite{oliehoek16}. 
In a finite horizon setting with length $T$, agent $i$ is trained with its policy $\pi^{i}$ to maximize a discounted accumulated reward at time-step $t$ computed as $J^{i}_{\pi} = \mathbb{E}_{\pi}\big[\sum_{\tau=0}^{T-t}\gamma^{\tau} r^{i}_{\tau+t} \big]$, where $\gamma$ denotes a discount factor for accumulating the rewards. 

There are two major challenges when we apply the MARL framework to the flexible coordination of agents in an industrial automated environment. 
\begin{itemize}
\item Long trajectory and extremely sparse reward (LTESR): an automated system is expected to maximize performance based on evaluation indices such as makespan and productivity estimated at the end of operations. Agents are, therefore, rewarded only at the last state transition even though they experience many state transitions to complete all tasks.
\item Asynchronized multi-agent (MA) decision making (AMADM): each agent, which corresponds to an automated process or robot, performs its tasks asynchronously. Most CTDE frameworks assume synchronized behaviors among agents for estimating joint state(-action) values, and few previous studies on asynchronous settings described in \S II.C can only be applied to off-policy RL algorithms with dense reward settings. 
\end{itemize}

\subsection{Proposal of MARL Framework for Industrial Automation}

\subsubsection{Foundational Framework}
to answer {\bf RQ1}, we propose a CTDE framework with a shared critic, termed CDSC: 
we adopt an actor-critic (AC) architecture for CDSC, where an actor policy $\pi$ and a critic $\hat{V}$ are modeled by different neural networks.    
Although widely referenced CTDE frameworks such as the MADDPG introduce individual joint critics that can be trained cooperatively with global state $s_{t}= \bigcup_{i=1}^{N} o^{i}_{t}$ and global reward $r_{t}^{g}$, they can only train networks based on state transitions caused by themselves under the AMADM settings. 
Thus, CDSC can provide more globalized training conditions for agents by aggregating all transition histories into a single critic.

\subsubsection{Algorithm Design}

\begin{algorithm}[!bp]
\caption{Episodic Training in CDSC}
\label{algo:marl_ppo}
\begin{algorithmic}[1]
\STATE Initialize policies $\{\pi^{i} \}_{i=1}^{N}$ parametrized by $\{\theta^{i}\}$, and shared critic $\hat{V}$ parameterized by $\phi$ 
\FOR{episode $=$1 to $N_{\mathrm{ep}}$}
\STATE Set rollout buffer $\mathsf{B} \leftarrow \emptyset $
\FOR{rollout $=1$ to $N_{\ell}$}
\STATE Set rollout $\mathsf{R} \leftarrow \emptyset$, and time $t=0$
\WHILE{${\it flag}=$ \textbf{True}} 
\FOR{Agent $i=1\textrm{ to }N$}
\IF{Agent $i$ needs to take action}
\STATE Make decision $a_{t}^{i} = \pi^{i}_{\theta} (a^{i}_{t}|o_t^{i})$ 
\ENDIF
\ENDFOR
\IF{$\{a_{t}^{i}\}\neq \emptyset$}
\STATE Execute actions $\{a_{t}^{i}\}$
\ENDIF
\STATE Count up $t$\,\texttt{+=}\,$1$
\FOR{Agent $i=1\textrm{ to }N$}
\IF{Agent $i$ took action at $t-1$}
\STATE Observe $o^{i}_{t},s_{t}$, and $r_{t}^{i}$
\STATE $\mathsf{R}$\,\texttt{+=}\,$[o^{i}_{t}, s_{t}, a^{i}_{t}, i, r^{i}_{t}, o^{i}_{t+1}, s_{t+1}]$
\ENDIF
\ENDFOR
\IF{All tasks have been completed}
\STATE \textit{flag}$=$ \textbf{False}
\ENDIF
\ENDWHILE
\STATE Compute $\hat{V}_{\tau}(s_{\tau})$, $\{C_{\tau}^{i}\}$, and $\{A^{i}_{\tau}\}$ by MC or GAE from $\tau=0$ to $t$ and add to ${\sf R}$
\STATE $\mathsf{B}\,\texttt{+=}\,\mathsf{R}$ 
\ENDFOR
\ENDFOR
\FOR{epoch $ = 1$ to $N_{k}$} 
\FOR{mini-batch $b = 1$ to $B$} 
\STATE Make mini-batch $b$ sampled from $\mathsf{B}$
\STATE Update $\{\theta^{i}\},\phi$ with data $b$
\ENDFOR
\ENDFOR
\end{algorithmic}
\end{algorithm}

to address the LTESR issue, we used an on-policy AC algorithm in the proposed MARL framework: 
The LTESR makes state values, estimated as TD errors frequently used in off-policy RL algorithms, uncertain because of many non-rewarded state transitions. 
We, therefore, used the on-policy algorithm in our proposed MARL frameworks to estimate state values as reward-to-go using the Monte Carlo (MC) method. 
As a SOTA RL algorithm, proximal policy optimization (PPO), which can train the policy of an agent conservatively, was implemented in our framework \cite{schulman17}. 
PPO also achieves good performance in the MA domain when implemented for CTDE (multi-agent PPO, MAPPO) \cite{yu21}. 
Hence, we applied MAPPO to CDSC, whereas the effectiveness of the PPO-based MARL framework is not well verified for our unique problem settings explained in \S III.A. 

A pseudocode for training agents using the proposed MARL framework is shown in Algorithm \ref{algo:marl_ppo}. 
Because industrial operations frequently involve batch processes (for instance, in logistics warehouses, shipping orders received from customers are processed hourly in batches), we assume an episodic training scheme for the proposed MARL framework. 
In this scheme, agent-environment interactions continue until all agents have completed all tasks (Algorithm 1, lines 22--24). 
To address the AMADM issue, an agent makes a decision and stores the state transition in the rollout buffer only when an action is required (lines 7--11 and 16--21). 
The PPO-based policy is trained conservatively by maximizing the following clipped objective function: 
\begin{eqnarray}
L^{i} (\theta^{i}) = \E_{b} \mleft[\min \mleft( \rho^{i}_{t}(\theta^{i}) A^{i}_{t}, 
\mathrm{clip} \mleft( \rho^{i}_{t}(\theta^{i}) , 1 \pm \epsilon \mright) A^{i}_{t} \mright) \mright] \ , \label{ppo_clip_loss}
\end{eqnarray}
where $\rho^{i}_{t}(\theta^{i})=\pi^{i}_\theta (a^{i}_{t} | o^{i}_{t})/\pi^{i}_{\theta_{\mathrm{old}}} (a^{i}_{t} | o^{i}_{t}) $ is a policy ratio with the old policy prior to the update parametrized by $\theta^{i}_{\mathrm{old}}$, $A^{i}_t$ is an advantage, and $\epsilon$ is the clipping range of the policy (line 33). 
Here, the expectation $\E_{b}[\ast]$ indicates the empirical average over a finite batch $b$ of samples. 
The advantage can be estimated using the MC method as $A^{i}_{t} = C^{i}_{t} - \hat{V}_{\phi}(s_{t}, i_{t})$, 
where $C_{t} = \sum_{\tau=0}^{T-t}\gamma^{\tau} r_{\tau+t} $ denotes a reward-to-go from time $t$ to the horizon $T$ when all agents have completed their tasks. 
The CDSC critic $\hat{V}$ takes as inputs not only state variables but also the ID of the agent $i$ to identify which agent's action contributes a corresponding state transition (line 26). 
The critic was trained in the following supervised learning manner as $L (\phi) = \E_{b} [ (\hat{V}_\phi - C_{t} )^{2} ]$.

\subsubsection{Generalized Advantage Estimation for Asynchronous Heterogeneous Multi-agent Settings}

the feature of a CDSC framework is that each agent can be trained with a shared critic aggregating all state transitions caused by different agents in a given rollout. 
Thus, the installation of a shared critic brings alternative methods of advantage estimation based on a TD error. 
Furthermore, generalized advantage estimation (GAE), which adjusts the bias-variance tradeoff of policy gradient estimates between MC and TD methods, can be used for computing $A^{i}_{t}$ in CDSC. 
Although a truncated version of GAE has already been proposed for episodic training by Schulman {\it et al}. \cite{schulman17}, we need to modify it to address the AMADM issue. 
Because the decision making of asynchronous agents would have irregular time intervals, GAE for AMADM in a finite episode is calculated as follows:
\begin{eqnarray}
\label{eq:gae}
A^{i}_{t} = \delta_{t} + (\gamma\lambda)^{\Delta t} \delta_{t+1} + \cdots + (\gamma\lambda)^{\Delta (T-1)} \delta_{T-1} \ , 
\end{eqnarray}
where $\lambda$ denotes a bias-variance tradeoff parameter [0,1], $T$ represents the episode length, $\Delta t$ denotes a time difference between $t$ and $t+1$, and $\delta_{t}$ denotes the following modified TD error: 
\begin{eqnarray}
\label{eq:td_error}
\delta_{t} = r_{t} + \gamma^{\Delta t}\hat{V}(s_{t+1}, i_{t+1}) - \hat{V}(s_{t}, i_{t}) \ . 
\end{eqnarray}
Compared with the original one, we can apply the proposed GAE to AMADM by simply extending the discounting coefficients $(\gamma\lambda)$ and $\gamma$ in Eqs. (\ref{eq:gae}) and (\ref{eq:td_error}) to an irregular interval setting. 
In addition, we can use the proposed GAE in the case of the AMADM of heterogeneous agents because the input of the value function includes the agent ID $i$. 
We use both the MC method and GAE to estimate the advantage in CDSC to maximize performance.

\subsection{Other MARL Frameworks for Comparison}

To answer {\bf RQ2} and {\bf RQ3}, we further introduced three different MARL frameworks to compare their coordination performance with that of CDSC.  
Table \ref{table:marl_framework} summarizes the features of critics in these frameworks, including CDSC from the viewpoint of information globalization. 
These frameworks assume the same functional policies $\{ \pi^{i} (a^{i}_{t} | o^{i}_{t}) \}_{i=1}^{N}$ for decentralized execution. 
Detailed training conditions of agents in the proposed MARL frameworks are described as follows: 
\subsubsection{IL with Local Reward (ILLR)}
an IL-based framework, where individual critic $V^{i}$ using the local state of an agent $o^{i}_{t}$ as input is trained with the local reward $r_{t}^{i}$ that can be estimated by itself. 
Because all information required for agent training is based on local observations, ILLR is the most localized training framework. 
A policy trained with ILLR corresponds to a short-sighted strategy, where robots are controlled to complete their tasks at hand as quickly as possible such as first-in-first-out and greedy heuristic algorithms \cite{ascheuer99,framinan08}. 

\subsubsection{IL with Global Reward (ILGR)}
an alternative IL-based framework with the same learning structure as ILLR, but the critic is trained with the global reward $r_{t}^{g}$ shared by all agents. 
Introduction of the global reward to the MAS is a typical cooperative setting in MARL \cite{oroojlooyjadid19}. 

\subsubsection{CTDE with Individual Critic (CDIC)}
a CTDE-based framework, where the individual critic $V^{i}$ can access information about the global state $s_{t}$. 
All agents are also trained with $r_{t}^{g}$ to promote coordination. 

We applied independent PPO (IPPO) to ILLR and ILGR and MAPPO to CDIC, respectively \cite{witt20,yu21}. 
Compared with these three frameworks, CDSC uses the most global information (data) for agent training. 
By comparing the training performance of these frameworks, we can determine which global information is effective for achieving coordination in industrial settings. 

\begin{table}[b]
\caption{MARL Frameworks with Different Critics}
\label{table:marl_framework}
\centering
\begin{tabular}{c|cccc}
\bhline{1pt} 
\multirow{2}{*}{Critic feature} & \multicolumn{4}{c}{MARL framework} \\
\cline{2-5}
& ILLR & ILGC & CDIC & CDSC \\
\bhline{1pt} 
State & \multicolumn{2}{c}{\cellcolor{red!30} Local} & \multicolumn{2}{c}{\cellcolor{blue!30} Global} \\
\hline
Reward & \cellcolor{red!30} Local & \multicolumn{3}{c}{\cellcolor{blue!30} Global} \\
\hline
Architecture & \multicolumn{3}{c}{ \cellcolor{red!30} Individual $(N)$} & \cellcolor{blue!30} Shared $(1)$ \\ 
\hline
State value & Localized & \multicolumn{2}{c}{$\Longleftrightarrow$} & Globalized \\
\bhline{1pt} 
\end{tabular}
\end{table}

\section{EVALUATION}
To answer {\bf RQs}, we applied the proposed MARL frameworks to the following automated order-picking simulator and evaluated the performance of trained policies. 

\subsection{Simulation Environment}
\subsubsection{Fully Automated Order Picking}
\begin{figure}[b]
\centering
\includegraphics[scale=0.43, angle=0]{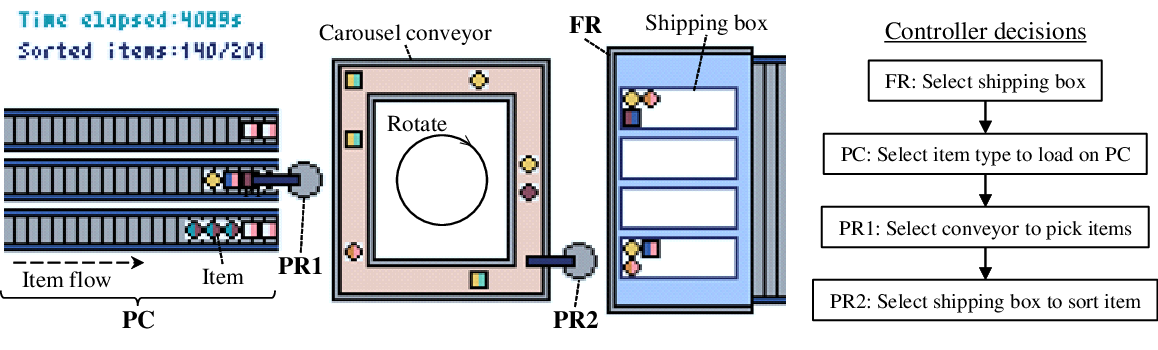}
\caption{Simulation environment. Left side: overview of AOPE composed of four automated processes (FR, PC, PR1, and PR2). Items with different shapes (square and circle) and colors denote different item types. 
Right side: decisions taken by the controller of each process.}
\label{fig_eval_env}
\end{figure}

the simulation of an automated order picking environment (AOPE) in a logistics warehouse was performed for quantitative evaluation of the proposed method. 
The overview of the AOPE is illustrated on the left side of Fig. \ref{fig_eval_env}. 
Ordered items are arranged from left to right in the figure. 
The AOPE consists of four types of MHE whose controller makes decisions for task selection, as shown on the right side of Fig. \ref{fig_eval_env}. 
The first one is a flow rack (FR) in which sorting boxes are allocated. 
Each sorting box corresponds to a shipping destination. 
When all ordered items are sorted into a box, the box is transported to the next working area, and FR replaces it with a new one. 
The FR controller selects a shipping box to start order picking tasks for its required items. 
The second one is a set of parallel conveyors (PC) that transports items from the inventory area to allocated shipping boxes in FR. 
There are three conveyors in parallel, and items are loaded from the inventory area by type. 
Each conveyor has six loading ports. 
The PC controller selects an item type to be loaded on the conveyor from the inventory area. 
The third one is a picking robot (PR1) that picks up items from the PC and places them on a carousel conveyor one by one. 
PR1 can also move among the conveyors to pick items from each conveyor. 
The PR1 controller selects a conveyor to pick items: conveyor selection occurs whenever the picking of the same item type is completed. 
The last one is another picking robot (PR2) that picks up items from a carousel conveyor and sorts them into shipping boxes one by one. 
The PR2 controller selects an allocated shipping box to sort items from the carousel conveyor. 
To prevent the simulation from being fixed to one work scenario, the item loading and replacement times of PCs and the picking time of both PR1 and PR2 are given normal distribution variations. 
Similar configurations were designed as a type of parts-to-picker systems \cite{xie21,suemitsu22}, whereas the AOPE was simplified to complete the simulation quickly as a training environment.

\subsubsection{Picking Order Data}

we evaluated the performance of the proposed MARL frameworks with two different picking order datasets in warehouse operations. 
The characteristics of these datasets are summarized in Table \ref{table:order_data}. 
The ``Orders'' in the table shows the number of unique sets of item types (``Types'') and shipping boxes (``Shippings''). 
In logistics warehouses, items are typically stored by type \cite{Bartholdi17}. 
Thus, the more item types in a picking order, the more item transfers there are in the upstream process, and the more difficult it is to optimize operations. 
To investigate the performance of our proposed algorithm for different picking orders, the item types are different, but items and shipping boxes are nearly the same. 
These picking orders have reasonable characteristics compared with previous studies in terms of the number of orders per type \cite{fussler19,xie21}. 

\begin{table}[t]
\caption{Picking order data}
\label{table:order_data}
\centering
\begin{tabular}{ccccc}
\bhline{1pt} 
Picking order & Orders & Items & Types & Shippings \\
\hline \hline
Low Mixed (LM) & 179 & 201 & 16 & 42 \\ 
High Mixed (HM) & 186 & 200 & 95 & 41 \\ 
\bhline{1pt} 
\end{tabular}
\end{table}

\subsection{RL Settings}
\subsubsection{Training Condition}
the proposed MARL frameworks were applied to train the four process controllers in an AOPE. 
Each process controller was regarded as an RL agent and trained in the advantage AC manner \cite{mnih16}. 
Table \ref{table:learnig_param} summarizes the hyperparameters used in this evaluation. 

\begin{table}[b]
\caption{Hyperparameters in MARL Settings}
\label{table:learnig_param}
\centering
\begin{tabular}{clc}
\bhline{1pt} 
Setting & Hyperparameter & Value \\
\hline \hline 
\multirow{7}{*}{\makecell{IPPO,\\ MAPPO}} & Clipping range $\epsilon$ & 0.2 \\ 
& Discount factor $\gamma$ & 0.99 \\ 
& Scaling factor $\zeta$ & 800 \\ 
& Rollouts $N_{\ell}$ & 64 \\
& Epochs $N_{k}$ & 5 \\
& Episodes $N_{ep}$ & 5000 \\
& Minibatch size & 64 \\
\hline
\multirow{4}{*}{Network} & Network & MLP \\
& Hidden layers & 2 \\
& Hidden units & 128 /layer \\
& Activation & Tanh() \\
\hline
\multirow{4}{*}{Optimizer} & Optimizer & Adam \cite{kingma15} \\ 
& Learning rates & $ 0.001 (\theta), 0.0003 (\phi)$ \\
& Decay rate & $0.8/250$ episodes \\
& Initialization & Orthogonal \\
\hline
\multirow{1}{*}{Reward} & $t_{\it ofs}$ & 6.6(LM), 6.4(HM) \\
\bhline{1pt} 
\end{tabular}
\end{table}

\subsubsection{Agent States}
the states of all agents are summarized in Table \ref{table:agents_states}. 
The number in brackets represents the number of states multiplied by the number of components.
As described in \S III, all agents were assumed to make decisions in a decentralized manner.

\begin{table}[tb]
\caption{States of Agents}
\label{table:agents_states}
\centering
\begin{tabular}{cp{6.6cm}}
\bhline{1pt} 
Agent & States \\
\hline
\hline
\multirow{1}{*}[-5ex]{FR} 
& Numbers of unsorted items and types in allocated shipping boxes (2), 
numbers of items and types waiting to be loaded on PC in allocated shipping boxes (2), 
number of unallocated shipping boxes (1), and
numbers of items and types in unallocated shipping boxes (2) \\ 
\hline
\multirow{1}{*}[-6ex]{PC} 
& ID of conveyor onto which items are loaded (1),
numbers of loaded items (1$\times$3), 
locations of the first and last loaded items (2$\times$3), 
numbers of items and types waiting to be loaded at loading ports (2$\times$6), and 
numbers of items loading onto conveyors and location of work-in-progress loading ports (2$\times$3) \\
\hline
\multirow{1}{*}[-9ex]{PR1} 
& Location and direction of PR1 (2),
numbers of items on carousel conveyor (1), 
numbers of items on PC (1$\times$3), 
locations of the first items on PC and the numbers of items whose types are the same as the first ones (2$\times$3), 
location of the last item on PC (1$\times3$), 
numbers of items waiting to be loaded at loading ports (1$\times$6), and 
numbers of items loading into conveyors and location of work-in-progress loading ports (2$\times$3) \\
\hline
\multirow{1}{*}[-4ex]{PR2} 
& Location and direction of PR2 (2), 
number of sorted items (1), 
array of items on carousel conveyor whether they can or cannot be sorted into shipping boxes ($28\times4$), and 
number of unsorted items of shipping boxes (1$\times$4) \\
\hline
Common 
& Time-step (1) and action mask (number of actions) \\
\bhline{1pt} 
\end{tabular}
\end{table}

\subsubsection{Action Mask}
we introduced an action mask to eliminate invalid actions at every decision making \cite{huang20}. 
The masked stochastic policy $\hat{\pi}^{i}_{\theta}$ computes the probability of a valid action $a^{i,k}_{t}$ as follows: 
\begin{eqnarray}
\hat{\pi}^{i}_{\theta} (a^{i,k}_{t}|o^{i}_{t}) = \pi^{i}_{\theta}(a^{i,k}_{t}|o^{i}_{t}) / \mbox{\small $\sum \nolimits_{a^{i,j}_{t}\in \mathcal{A}^{i}} $} m(a^{i,j}_{t}) \pi^{i}_{\theta} (a^{i,j}_{t}|o^{i}_{t}) , 
\label{eq:action_mask}
\end{eqnarray}
where $m(a^{i,j}_{t})$ denotes the action mask of an  agent $i$ that outputs 0/1 when the $j$-th action is valid/invalid at $t$. 
The action mask reduces $\rho^{i}_{t}(\theta^{i})$ and stabilize gradient updates.

\subsubsection{Reward Design}
the objective of agents in AOPE was to minimize the makespan $T_{c}$ of all order picking tasks. 
Thus, the local reward $r^{i}_{t}$ for ILLR was set to the negative value of the elapsed time until an agent selects the next task: the RL policy was expected to learn to complete the current task as quickly as possible.
Agents in ILLR received $r^{i}_{t}$ as an immediate reward for every decision they made.
In contrast, agents in the other three MARL frameworks were trained in the LTESR setting: the global reward was only given at the terminal state transition. 
The terminal global reward $r^{g}_{tml}$ was estimated as follows:
\begin{eqnarray}
r^{g}_{tml} = 
\begin{cases}
{2 \cdot [(t^{\rm ks}_{c}-t_{\it{ofs}})^{4} - 1] \ \ \ \ \text{if $t_{c}\geq t_{\it{ofs}}$}} \\
{-2 \cdot [(t^{\rm ks}_{c}-t_{\it{ofs}})^{4} + 1]\ \ \text{otherwise}} 
\end{cases}
,
\label{eq:global_reward}
\end{eqnarray}
where $t^{\rm ks}_{c}$ denotes the ksec-unit $T_{c}$ when the last item is sorted into the last shipping box, and $t_{\it{ofs}}$ denotes an offset depending on the picking order dataset. 
Because $t_{c}$ represents a ksec-order value and the AOPE was computed at 10 fps, agents exhibit a long trajectory through the simulation. 
Therefore, if a well-known $\gamma \sim 0.99$ is used in the training, the impact of $r^{g}_{tml}$ on $C_{t}$ will be significantly decayed because of the long trajectory. 
To propagate $r^{g}_{tml}$ through the trajectory, we introduced a scaling factor $\zeta$ listed in Table \ref{table:learnig_param} into the discounting coefficients in Eqs. (\ref{eq:gae}) and (\ref{eq:td_error}) as $(\gamma\lambda)^{\Delta t/\zeta}$ and $\gamma^{\Delta t/\zeta}$.

\subsection{Rule-based Control}

To compare the performance of MARL-based control, we established a set of control rules for each process controller. 
Control features in AOPE are summarized as follows:
\begin{itemize} 
\item Decentralized control decisions with hierarchical structure (FR$\rightarrow$PC$\rightarrow$PR1$\rightarrow$PR2). 
\item Task execution can be parallelized in a process (processes in FR and PC). 
\item Each process has different task granularities (shipping destinations for FR, item types for PC, and items for PR1 and PR2). 
\item Each process contains stochastic uncertainties (e.g., picking speeds of PR1 and PR2).
\end{itemize}
We have found that there is no prior approach using mathematical optimizations, metaheuristics, or artificial intelligence that can comprehensively handle the above four features even though recent studies have covered some of them \cite{rasmi22,kordos20,park21,suemitsu22}. 
Thus, the rule-based control, still actively researched for warehouse optimization, was used for our evaluation \cite{bozer18}. 
The sets of control rules of all processes are summarized in Table \ref{table:control_rule}, where a number in parentheses indicates the number of control rules for each process. 
We applied a seed algorithm to four of the eight rules of FR \cite{elsayed81}. 
The seed algorithm can select a shipping box with the best score obtained by comparing items in candidate and allocated boxes. 
In this evaluation, we estimated the score of shipping boxes as the similarity of allocated boxes. 
The similarity was calculated by accumulating weights $(w_{s}, w_{d})$, where $w_{s}$ is added to the score if one item type in the candidate box is required for the allocated boxes, whereas $w_{d}$ is added if the item type is excluded from the allocated boxes. 
To expand the choices of shipping boxes, we established seed algorithm-based rules by changing weights, as $(w_{s}, w_{d})=(1,0),(0,1),(1,-1)$, and $(-1,1)$.

\begin{table}[!t]
\centering
\caption{Control rules of four processes}
\label{table:control_rule}
\begin{tabular}{cl}
\bhline{1pt} 
Process & Control rules for task selection \\
\hline \hline
\multirow{3}{*}{FR} & A shipping box with the most/fewest items (2) \\
& A shipping box with the most/fewest item types (2) \\
& A shipping box selected by seed algorithm (4) \\
\hline
\multirow{2}{*}{PC} & An item type with the most/fewest items, or \\
& \ farthest/closest to PR1 (8) \\
\hline
\multirow{2}{*}{PR1} & A conveyor loading a head item type with the \\
& \ most/fewest items, or farthest/closest to PR1 (8) \\
\hline
\multirow{2}{*}{PR2} & A shipping box with the most/fewest unsorted \\
& \ items, or farthest/closest to PR2 (8) \\
\bhline{1pt} 
\end{tabular}
\end{table}

\subsection{Results and Discussions}

\subsubsection{Improving Training Performance with GAE in CDSC}
\begin{figure}[!t]
\centering
\includegraphics[scale=0.17, angle=0]{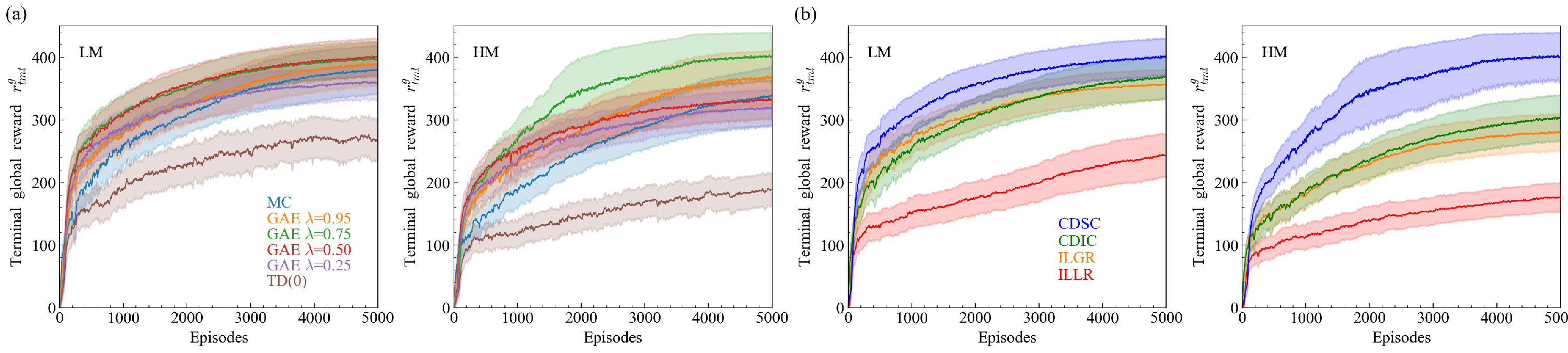}
\includegraphics[scale=0.17, angle=0]{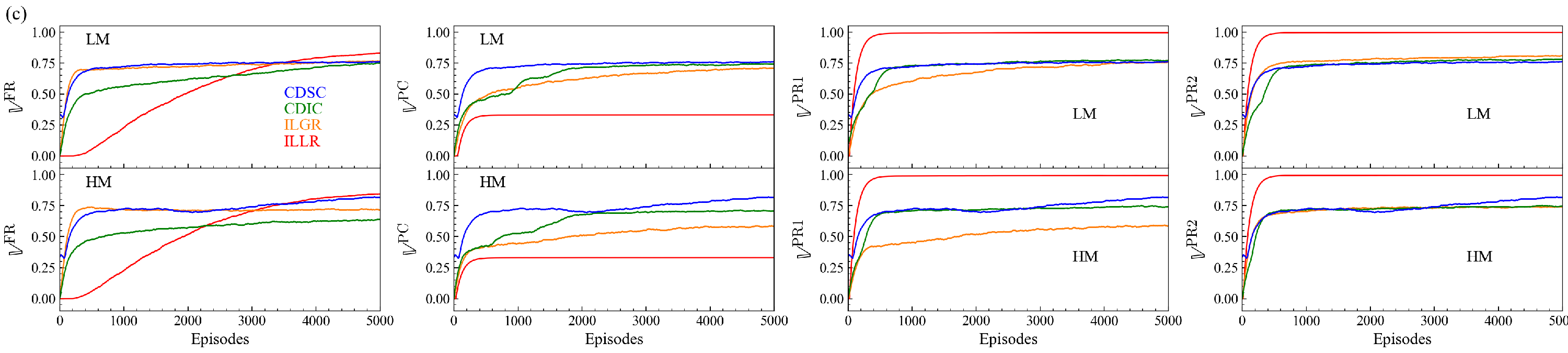}
\caption{Training results of proposed MARL frameworks in AOPE with LM and HM picking orders. (a) learning curves of CDSC with different advantage estimations. (b) Comparison of learning curves among four MARL frameworks. (c) Comparison of explained variances of each AOPE agent among four MARL frameworks.}
\label{fig_learning_curve}
\end{figure}

Figure \ref{fig_learning_curve} (a) shows training curves of CDSC with different advantage estimations in AOPE with LM and HM picking orders. 
Each curve shows averaged performance with a standard deviation over three random seeds. 
TD(0) has the worst performance among all methods, suggesting the difficulty of applying the TD method in LTESR as explained in \S III.B.2. 
Compared with the MC method, GAE accelerated the training and achieved significantly better performance in the early phase. 
This performance superiority was maintained until the end of training for $\lambda=0.5\sim0.95$. 
Hence, the proposed GAE is an effective advantage estimation for CDSC in AOPE. 

\subsubsection{Comparison of Different Control Methods}
we summarize the training results of four MARL frameworks in Fig. \ref{fig_learning_curve} (b) and simulation results of AOPE obtained from different control methods in Table \ref{table:performance_comparison}. 
Each value shows an average and standard deviation of $T_{c}$ sampled from 192 rollouts. 
As the baseline of each simulation, we added the results, where all controllers selected their tasks at random, as ``Random choice.'' 
The performance of rule-based control represents the result of the best combination rules (4096 in total) listed in Table \ref{table:control_rule}. 
ILLR, the most localized MARL framework, achieved comparative performance to the rule-based control for the LM picking order, outperforming the HM one. 
Because these methods control robots with a short-sighted plan to complete their tasks at hand as quickly as possible, the advantage of ILLR over rule-based control for the HM picking order can be attributed to the flexibility in task selection. 
The ILLR-trained policy can flexibly select a task depending on the input $o_{t}^{i}$ reflecting the operation status compared with the rule-based policy whose selection is solely based on the implemented rule. 
Such flexibility is more effective for the HM picking order whose task selection in PCs is more frequent due to the large variety of item types. 
Thus, the MARL-based control can provide more efficient operations in industrial automated systems than the rule-based control, even with localized training.

\begin{table}[!b]
\caption{Comparison of Makespans among Different Control Methods}
\label{table:performance_comparison}
\centering
\begin{tabular}{cc|cc}
\bhline{1pt} 
\multicolumn{2}{c|}{\multirow{2}{*}{Control method}} & \multicolumn{2}{c}{Picking order} \\
\cline{3-4}
& & LM & HM \\
\hline \hline
\multicolumn{2}{c|}{Random choice} & 5385.0$\pm$188.9 s & 5227.8$\pm$180.6 s \\ 
\hline 
\multicolumn{2}{c|}{Rule-based} & 3273.4$\pm$110.6 s & 3599.5$\pm$120.1 s \\ 
\hline 
\multirow{5}{*}{MARL} & \multicolumn{1}{|c|}{ILLR} & 3278.2$\pm$120.2 s & 3334.2$\pm$103.9 s \\
\cline{2-4}
& \multicolumn{1}{|c|}{ILGR} & 2941.9$\pm$62.9 s & 2975.9$\pm$90.8 s \\
\cline{2-4}
& \multicolumn{1}{|c|}{CDIC} & 2914.9$\pm$87.5 s & 2891.3$\pm$102.8 s \\
\cline{2-4}
& \multicolumn{1}{|c|}{CDSC (MC)} & 2884.7$\pm$94.9 s & 2793.1$\pm$123.2 s \\ 
\cline{2-4}
& \multicolumn{1}{|c|}{CDSC (GAE)} & {\bf2834.3$\pm$69.4 s} & {\bf2638.2$\pm$101.3 s} \\ 
\bhline{1pt} 
\end{tabular}
\end{table}

We answer the research questions as follows.

{\bf RQ1:} 
as shown in Table \ref{table:performance_comparison}, CDSC-based control with GAE ($\lambda=0.5,0.75$ for LM, HM picking orders) achieved the shortest makespans among all methods. 
Thus, the MARL agents achieve coordination by introducing both the globalization of training information and the unification of state transitions to the shared critic. 

{\bf RQ2:} 
the most effective globalization from ILLR is the reward setting caused by ILGR, where agents share the global reward. 
Figure \ref{fig_learning_curve} (c) shows smoothed explained variance of state values averaged over different trials computed as $ \mathbb{V}^{i} = 1 - \mathrm{Var}(V^{i}_{\mathrm{exp}}-V^{i}_{\phi})/\mathrm{Var}(V^{i}_{\phi})$, where $V^{i}_{\mathrm{exp}}$ denotes a true state value obtained from experiments, and $V^{i}_{\phi}$ denotes the prediction by the critic. 
$\mathbb{V}^{\mathrm{PR_{1}}}$ and $\mathbb{V}^{\mathrm{PR_{2}}}$ of ILLR rapidly converged to 1 and almost fully predicted the actual state values compared with the other three frameworks, 
whereas $\mathbb{V}^{\mathrm{FR}}$ and $\mathbb{V}^{\mathrm{PC}}$ of ILLR yield poor prediction. 
This result suggests that ILLR agents were trained so locally that the agents in downstream processes could easily infer individual environmental changes, whereas ones in upstream processes, significantly affected by downstream performance, could not achieve reasonable prediction. 
This localized training tendency was significantly eliminated by introducing the global rewards: 
$\mathbb{V}^{\mathrm{FR}}$ and $\mathbb{V}^{\mathrm{PC}}$ were significantly improved by ILGR in exchange for a slight decrease in $\mathbb{V}^{\mathrm{PR_{1}}}$ and $\mathbb{V}^{\mathrm{PR_{2}}}$.
The relatively marginal contribution of state globalization via CDIC may be attributed to the difficulty in critic training due to the increase in the number of input states, as listed in Table \ref{table:agents_states}.
In support of this consideration, CDSC with the MC method and the same advantage estimation outperformed CDIC. 
Although the representation ability of CDIC individual critics was lost, this drawback may have been resolved by a CDSC single critic. 
Furthermore, the performance of CDSC was improved by the proposed GAE, and thus, it achieved the shortest $T_{c}$ among all control methods. 

\begin{figure}[t]
\centering
\includegraphics[scale=0.37, angle=0]{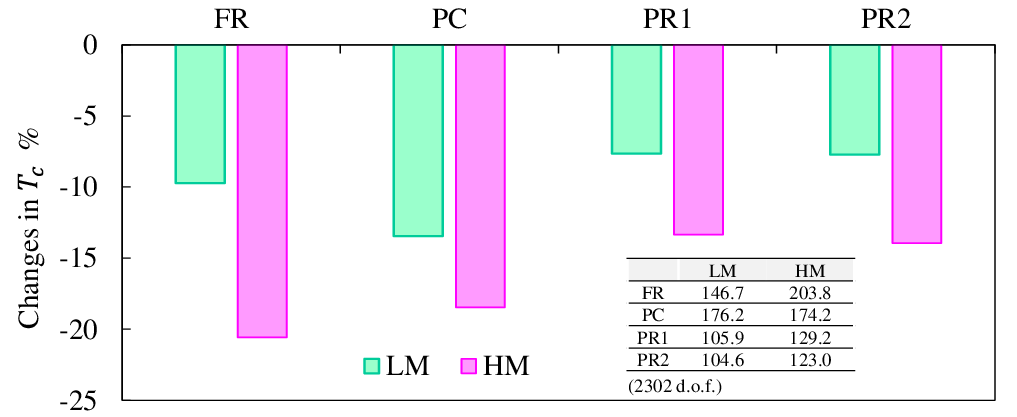}
\caption{Percentage change in makespan when switching from CDSC-trained to ILLR-trained policies for each agent. Statistics of Welch's $t$-test between CDSC and switched results are listed in the inserted table: all corresponding $p$-values satisfy $p<0.001$ (95\% confidence interval).}
\label{fig_comp_illr_cdsc}
\end{figure}

{\bf RQ3:} 
we evaluated performance degradation caused by changing the policy for each process from CDSC to ILLR. 
Figure \ref{fig_comp_illr_cdsc} shows the percentage change in the makespan when switching from CDSC-trained to ILLR-trained policies for each agent. 
The performance of CDSC was degraded even if its downstream policies (namely, PR1 and PR2) were replaced with ILLR-trained policies. 
The performance degradation in downstream processes suggests the coordination throughout the overall processes, and such overall coordination may be the first benefit of overall optimization via CDSC. 
Furthermore, the ILLR-trained policy of FR (and PC) results in significant performance degradation. 
This trend can be seen more prominently with the HM picking order, where PCs make decisions more frequently due to many item types. 
Such results are consistent with the comparison results of explained variances, where ILLR causes poor state value predictions in upstream processes, as described above. 
The upstream processes make decisions more sparsely because their task granularity is bulkier, such as shipping box (FR) and item type (PC). 
This sparse decision making is more susceptible to environmental changes caused by the downstream processes; thus, the ILLR framework fails to optimize the control of operations. 
Hence, the second benefit of overall optimization in warehouse automation via CDSC may be the improvement in efficiency in upstream processes.

\section{CONCLUSION}
To clarify the impact of overall optimization on industrial automation, we explored an efficient MARL framework that enables practical robot coordination. 
In the proposed framework, agents were trained in CDSC, a CTDE manner using both globalized rewards and single shared critic. 
Furthermore, we proposed the modified GAE for policy update to improve the performance of CDSC to address the two major issues in typical industrial settings: LTESR and AMADM. 
The evaluation results show that the CDSC-based control applied to task selections of MHE in AOPE can achieve the shortest makespan compared with other MARL frameworks and rule-based controls. 
The results also suggest that the overall optimization has the following advantages for warehouse automation: bottom-up efficiency through process coordination and further efficiency of upstream process control. 
Although the obtained knowledge is limited to our experimental design, we believe that we can qualitatively and quantitatively clarify the impact of overall optimization on various types of automated systems with different layouts and configurations by introducing our proposed MARL frameworks for future work.


\bibliographystyle{IEEEtran}
\bibliography{mybibfile}

\end{document}